\documentclass[letterpaper]{article} % DO NOT CHANGE THIS
\def\SOLNavDocumentClassLoaded{1}
\def\SOLNavFullVersion{1}
\ifdefined\SOLNavDocumentClassLoaded
\else
  \documentclass[letterpaper]{article} % DO NOT CHANGE THIS
\fi
\ifdefined\SOLNavFullVersion
  \usepackage[preprint]{aaai2027}  % DO NOT CHANGE THIS
\else
  \usepackage[submission]{aaai2027}  % DO NOT CHANGE THIS
\fi
\usepackage[hyphens]{url}  % DO NOT CHANGE THIS
\usepackage{graphicx} % DO NOT CHANGE THIS
\usepackage{natbib}  % DO NOT CHANGE THIS AND DO NOT ADD ANY OPTIONS TO IT
\usepackage{caption} % DO NOT CHANGE THIS AND DO NOT ADD ANY OPTIONS TO IT
\usepackage{algorithm}
\usepackage{algpseudocode}
\usepackage{amsmath}
\usepackage{amssymb}
\usepackage{bm}
\usepackage{newfloat}
\usepackage{listings}
\DeclareCaptionStyle{ruled}{labelfont=normalfont,labelsep=colon,strut=off} % DO NOT CHANGE THIS
\floatstyle{ruled}
\newfloat{listing}{tb}{lst}{}
\floatname{listing}{Listing}
\usepackage{booktabs}
\usepackage{multirow}
\usepackage{color}
\usepackage{xcolor}

\title{SOL-Nav: Structured Observation Language for Efficient and Generalizable Vision-Language Navigation}

\ifdefined\SOLNavFullVersion
  \author {
      Daojie Peng\textsuperscript{\rm 1}\equalcontrib,
      Fulong Ma\textsuperscript{\rm 1}\equalcontrib,
      Jun Ma\textsuperscript{\rm 1,2}\thanks{Corresponding author.}
  }

  \affiliations {
      \textsuperscript{\rm 1}The Hong Kong University of Science and Technology (Guangzhou)\\
      \textsuperscript{\rm 2}The Hong Kong University of Science and Technology\\
      \{dpeng108, fmaaf\}@connect.hkust-gz.edu.cn, jun.ma@ust.hk
  }
\else
  \author{Anonymous Authors}
  \affiliations{Anonymous Affiliation}
\fi

\begin{document}

\maketitle

% ============================================================
% Abstract
% ============================================================
\begin{abstract}
 Vision-Language Navigation (VLN) requires an embodied agent to navigate complex environments by following natural language instructions, which typically demands tight fusion of visual and language modalities. Existing VLN methods often convert raw images into visual tokens or implicit features, requiring large-scale visual pre-training and suffering from poor generalization under environmental variations (e.g., lighting, texture). To address these issues, we propose SOL-Nav (Structured Observation Language for Navigation), a novel framework that translates egocentric visual observations into compact structured language descriptions for efficient and generalizable navigation. Specifically, we divide RGB-D images into a N×N grid, extract representative semantic, color, and depth information for each grid cell to form structured text, and concatenate this with the language instruction as pure language input to a pre-trained language model (PLM).
  Experimental results on standard VLN benchmarks (R2R, RxR) and real-world deployments demonstrate that SOL-Nav significantly reduces the model size and training data dependency, fully leverages the reasoning and representation capabilities of PLMs, and achieves strong generalization to unseen environments.
\end{abstract}

\begin{figure*}[h]
    \setlength{\abovecaptionskip}{0pt}
    \setlength{\belowcaptionskip}{0pt}
    \centering
    \includegraphics[width=1.0\linewidth]{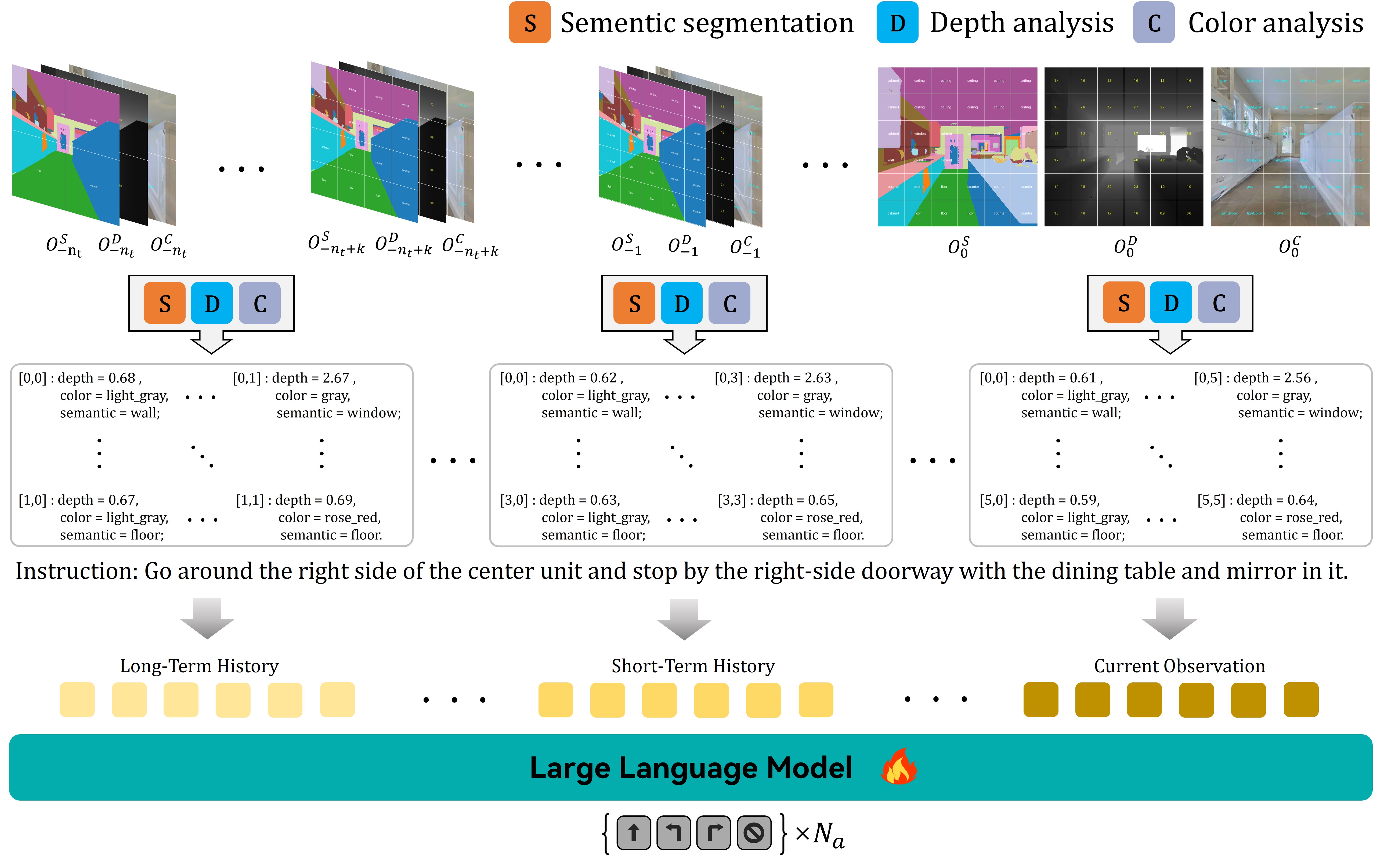}
    \captionsetup{font={small}}
    \caption{Pipeline of SOL-Nav. RGB-D observations are converted into structured textual descriptions with 2×2/4×4/6×6 multi-resolution grids (long/short-term history, current observation) encoding depth, semantic, and color information. The structured observation sequence, navigation instruction, and system description form a pure language prompt, which is input to a LLM to predict a consecutive action block for the agent.}
    \label{fig:architecture_new}
\end{figure*}

% ============================================================
% 1. Introduction
% ============================================================
\section{Introduction}
\label{sec:intro}

Vision-Language Navigation (VLN) is a core challenge in embodied AI, requiring an agent to understand natural language instructions and complete autonomous navigation in complex real or virtual environments. This task integrates computer vision, natural language processing, and spatial reasoning capabilities, and is a key step towards general embodied intelligence \cite{cheng2024navila, wu2024vision}. With the rapid development of embodied AI, research in the VLN field has gradually shifted from early supervised learning methods to multimodal fusion methods based on Large Language Models (LLMs), aiming to leverage the powerful reasoning capabilities of pre-trained large models to improve the performance and generalization of navigation agents \cite{lin2025navcot, goetting2024end,peng2025lovon}.

Existing VLN methods mostly convert image observations into language-like tokens for multimodal fusion: image encoders map pixel-level information to visual tokens, which are concatenated with language instruction tokens and fed into multimodal models for action prediction \cite{zhang2024uni, goetting2024end}. Despite achieving partial vision-language fusion, these methods have three critical limitations: 1) they rely on large-scale domain-specific training data and train visual representations from pixel-level features from scratch, leading to high training costs and slow convergence \cite{zhang2024uni}. 2) their generalization is limited by environmental changes (e.g., lighting, object texture, scene layout), resulting in poor performance in unseen environments \cite{cheng2024navila, goetting2024end}. 3) they fail to fully reuse pre-trained language model (PLM) capabilities due to a modal gap between visual and language tokens, hindering the exploitation of LLMs' strengths in reasoning and decision-making \cite{lin2025navcot}.

Recent LLM-based VLN methods attempt to address these issues by framing navigation as language reasoning tasks. For example, NavCoT decomposes navigation decisions into three interconnected reasoning steps: Future Imagination, Visual Information Filtering, and Action Prediction \cite{lin2025navcot}; Uni-NaVid outputs low-level robot control actions via a unified end-to-end framework \cite{zhang2024uni}; NaVILA improves generalization capability with multi-source data fusion \cite{cheng2024navila}. However, these methods still cannot fully solve the vision-language modal alignment problem: visual information processing still relies on image encoder feature extraction, and visual signals are not converted into structured text directly interpretable by LLMs, thus still requiring massive multimodal training data to bridge the modal gap.

To tackle this fundamental issue, we propose SOL-Nav, a VLN framework that directly converts RGB-D observations into structured textual representations, concatenates them with natural language instructions, and feeds the unified language input into a PLM for robot action prediction.  Specifically, Specifically, we first use pre-trained semantic segmentation models (e.g., SegFormer \cite{xie2021segformer}, YOLO-SAM \cite{li2025yolosam}, Grounded SAM \cite{ren2024grounded}.) to predict semantic regions from the input image, divide the image into a $N \times N$ grid, extract representative semantic categories, colors, and depth information for each grid cell to form structured textual observations. We then concatenate these structured observations with navigation instructions as pure language input to a PLM, which directly predicts the agent's navigation actions.

SOL-Nav has three key advantages: 1) \textbf{Reduced training cost}: Structured textual representation of visual information eliminates the need for scratch training of visual encoders, enabling full reuse of PLM capabilities and only requiring small-scale navigation data for fine-tuning; 2) \textbf{Stronger generalization}: Structured text avoids the impact of environmental factors (e.g., lighting, texture) on visual observations, allowing the model to adapt better to unseen environments; 3) \textbf{Simplified pipeline}: No complex multimodal encoders or fusion modules are needed, and navigation decisions are made via a pure language model, reducing model complexity and computational cost.

We evaluate SOL-Nav on two mainstream VLN datasets (R2R-CE \cite{anderson2018vision} RxR-CE \cite{ku2020room})and compare it with state-of-the-art (SOTA) methods (including Uni-NaVid, NaVILA, etc.). Experimental results show that SOL-Nav achieves SOTA or comparable performance on navigation success rate, path efficiency, and other metrics with a far smaller model. We also validate SOL-Nav's practicality via real-world robotic deployments.

The main contributions of this paper can be summarized as follows:

1. A novel VLN method that converts image observations into structured textual representations is proposed, which achieves seamless alignment between visual information and language models and fully reuses the reasoning capabilities of pre-trained large language models.

2. A grid-based visual key information extraction strategy is designed, which converts visual observations into textual forms that can be directly understood by language models through structured descriptions of semantics, colors, and depth information.

3. Comprehensive experimental verification is conducted on the R2R-CE and RxR-CE datasets as well as in the real-world, proving the advantages of the proposed method in reducing training resources, improving generalization capability, and navigation performance.

The rest of this paper is structured as follows: Section \ref{sec:related_works} reviews related work on VLN, embodied navigation datasets, and vision-language-action models; Section \ref{sec:methods} elaborates on the SOL-Nav framework, including problem formulation, structured observation construction, and model architecture; Section \ref{sec:experiments} presents experimental setups and results; Section \ref{sec:conclusion} concludes the work and outlines future directions.

% ============================================================
% 2. Related Work
% ============================================================
\section{Related Work}
\label{sec:related_works}

\subsection{Vision-Language Navigation}

Vision-Language Navigation (VLN) requires agents to execute long-horizon planning and precise movements based on natural language instructions. Early VLN studies adopted discrete topological graphs for node-to-node transitions \cite{anderson2018vision,ku2020room,qi2020reverie}, which simplifies the problem but evades real-world challenges such as obstacle avoidance and low-level path planning. To bridge the gap with real-world applications, Vision-and-Language Navigation in Continuous Environments (VLN-CE) \cite{savva2019habitat} was proposed, which requires agents to output low-level continuous control actions and has significantly improved navigation accuracy in simulated environments.  However, constrained by task-specific network architectures and limited training resources, existing models frequently encounter performance bottlenecks in Zero-shot Generalization and Sim-to-Real transfer.

Recent studies \cite{chen2024mapgpt,lin2025navcot,long2024instructnav,qiao2025open,zhang2025nava,zhou2024navgpt} have attempted to incorporate general foundation models to enhance the robustness of VLN in realistic scenarios. Nevertheless, in the absence of diverse fine-tuning data for downstream tasks, these models still struggle to fully align with the specific demands of the navigation domain. Motivated by this, we propose a novel framework based on the decoupling of high-level semantic representations. We observe that SOTA semantic segmentation models exhibit remarkable cross-domain generalization capabilities. If VLN models directly process raw sensor data, they are highly susceptible to overfitting the specific visual appearances of their training environments. Therefore, although SOL-Nav still leverages foundation model capabilities for decision-making, it abandons the traditional paradigm of directly inputting raw RGB-D data, instead using structured semantic textual features to improve generalization.

\subsection{Embodied Navigation Datasets} 
To support the training and evaluation of embodied navigation strategies, the research community has constructed numerous datasets and corresponding evaluation systems \cite{duan2022survey,liu2025aligning,mavrogiannis2023core,zhu2021deep}. These data resources hold foundational significance for the advancement of the navigation field. In the following, the relevant core datasets for the tasks involved in this study will be outlined. In the field of vision-language navigation, R2R-CE \cite{anderson2018vision} and RxR-CE \cite{ku2020room} have become widely adopted benchmarks, both providing human-annotated navigation instructions and corresponding ground-truth motion trajectories. For the object-goal navigation task, there are currently several evaluation benchmarks built on different simulation platforms, including HM3D \cite{ramakrishnan2021habitat}, MP3D \cite{chang2017matterport3d}, and Aithor \cite{zhu2017target}. 
In the field of Embodied Question Answering (EQA), datasets such as MP3D-EQA \cite{wijmans2019embodied}, MT-EQA \cite{yu2019multi}, Graph-EQA \cite{tan2023knowledge}, and MX-EQA \cite{islam2023eqa} each focus on different attributes of questions.

This study focuses on the Vision-Language Navigation (VLN) direction and conducts research using its continuous environment version, VLN-CE. VLN-CE requires agents to output continuous control actions, which more closely aligns with the actual motion scenarios of physical robots and is the commonly used standard for modern VLN research.

\subsection{Vision-Language-Action Model for Navigation}
Recently, leveraging multimodal large models as pretrained backbones for navigation tasks has emerged as a significant trend, with its core objective being to exploit the commonsense knowledge inherent in these models to enhance navigation capabilities. A typical implementation involves representing navigation actions as textual sequences, thereby framing the task as a next-token prediction problem within large language models (LLMs). 
Several notable studies, including the studies \cite{gao2025octonav,zhang2024navid,wang2025monodream,wei2025streamvln,zhang2024uni,zheng2024towards}, retain the discrete action space originally defined in VLN-CE. In these approaches, navigation decisions are made by designing task-specific vocabularies that function as categorical output labels for the language model.
In contrast, methods such as RoboPoint \cite{yuan2024robopoint} and NaviMaster \cite{luo2025navimaster} circumvent the limitations of discrete action spaces by formulating navigation as a pixel-grounding task. Nevertheless, these approaches still depend on supplementary modules for action execution, such as camera calibration and point-of-goal navigation policies.

Building on this, UniVLA \cite{bu2025univla} and TrackVLA \cite{wang2025trackvla} propose end-to-end frameworks that map LLM-extracted latent features directly to continuous trajectories executable by robotic platforms, streamlining the perception-to-control pipeline. 
While these methods leverage LLM comprehension and reasoning for complex navigation, they follow a monolithic end-to-end paradigm that maps raw sensor inputs directly to navigation commands. This leads to three critical issues: 1) high sensitivity to environmental variations and limited generalization; 2) reliance on massive training data for modal alignment; 3) dependence on large-scale models ($\geq$7B parameters), resulting in low inference speed and hindering real-world deployment on embodied systems.

To address these issues, SOL-Nav first preprocesses visual data to extract structured semantic, color, and depth information, which is then fed into a transformer model in textual form. This design isolates the navigation model from sensor and environmental disturbances, enhancing robustness. Furthermore, this preprocessing step enables the use of a small-scale model (<1B parameters) to achieve SOTA performance, which is highly advantageous for the practical deployment of embodied navigation models on physical robots.

% ============================================================
% 3. Method
% ============================================================
\section{Method}
\label{sec:methods}

% \subsection{Vision-Language Navigation Task Definition}
\subsection{Problem Formulation}

The VLN task requires an embodied agent to navigate in unknown environments following natural language instructions, integrating visual perception, language understanding, and spatial reasoning capabilities. Given a sequence of visual observations  $O = {O_{-n_t}, ..., O_{-1}, O_0}$  where  $O_t$  represents the visual observation at time step  $t$ , and a natural language task instruction  $I$, the agent is required to predict a sequence of navigation actions to complete the specified task.

Most existing methods represent $O$ with implicit visual features or vision-language tokens, which requires complex multimodal fusion and scratch visual encoder training. In contrast, we translate raw RGB-D observations into compact, structured natural language, enabling the navigation policy to be implemented purely in a PLM without any scratch visual encoder training.

% \subsection{Task Formulation in This Work}

In this work, we formulate the VLN task as an action chunk prediction problem, where the agent predicts a sequence of $N_a$ consecutive actions (action chunk) based on structured historical observations, current observation, and task instruction as illustrated in Figure \ref{fig:architecture_new}. The input and output of the model are as follows:

\noindent \textbf{Input}:
\begin{itemize}
    \item \textit{System Description $D_{\text{system}}$}: It serves as the foundational guidance for the language model, clearly defining the robot's executable action space, task objective, and decision-making constraints. 
    \item \textit{Current Structured Observation $O_0$}: A $N_{\text{curr}} \times N_{\text{curr}}$ grid-based structured observation matrix, where each grid cell element $[O_0]_{ij}$ (for row index $i$ and column index $j$, $1 \leq i,j \leq N_{\text{curr}}$) is formed by concatenating three types of string-based information at the corresponding position: 
    \begin{itemize}
        \item average depth value $o^d_{ij} \in O^D$ (represented as a string), 
        \item semantic category $o^s_{ij} \in O^S$ (the dominant semantic class in the grid), 
        \item representative color $o^c_{ij} \in O^C$ (mapped from HSV values to standard color names).
    \end{itemize}
    Mathematically, the element at position $(i,j)$ of $O_0$ is defined as:
    \begin{align}
        [O_0]_{ij} = o^d_{ij} \oplus o^s_{ij} \oplus o^c_{ij}
    \end{align}
    where $\oplus$ denotes the string concatenation operation, and $O^D, O^S, O^C$ are all $N_{\text{curr}} \times N_{\text{curr}}$ string matrices corresponding to depth, semantic, and color information respectively.
    \item \textit{Compressed Short-Long Historical Observations $\hat{O}_{-n_t}, \dots, \hat{O}_{-1}$}: Given $n_t$ past frames, we categorize them into $n_{\text{short}}$ short-term and $n_{\text{long}}$ long-term memories (satisfying $n_t = n_{\text{short}} + n_{\text{long}}$). Each category adopts a distinct grid resolution: $N_{\text{short}} \times N_{\text{short}}$ for short-term memories and $N_{\text{long}} \times N_{\text{long}}$ for long-term memories. For any historical observation matrix $\hat{O}_{-k}$ ($1 \leq k \leq n_t$), its grid cell elements follow the same concatenation rule as $O_0$: $\hat{o}^s_{ij} \oplus \hat{o}^c_{ij} \oplus \hat{o}^d_{ij}$, retaining consistency in the semantic-depth-color representation.
    \item \textit{Task Instruction $I_{\text{task}}$}: A natural language instruction describing the navigation task.
\end{itemize}

\noindent \textbf{Output}: 
An action block $A = [a_1, a_2, ... , a_{N_a}]$, where each $a_i \in {0, 1, 2, 3}$ represents a navigation action: 0 for stop, 1 for turn left 15 degrees, 2 for turn right 15 degrees, and 3 for move forward 25 cm.

\subsection{Structured Observation Language Prompt Construction}\label{sec:model_setup}

\begin{figure}
    \centering
    \includegraphics[width=1.0\linewidth]{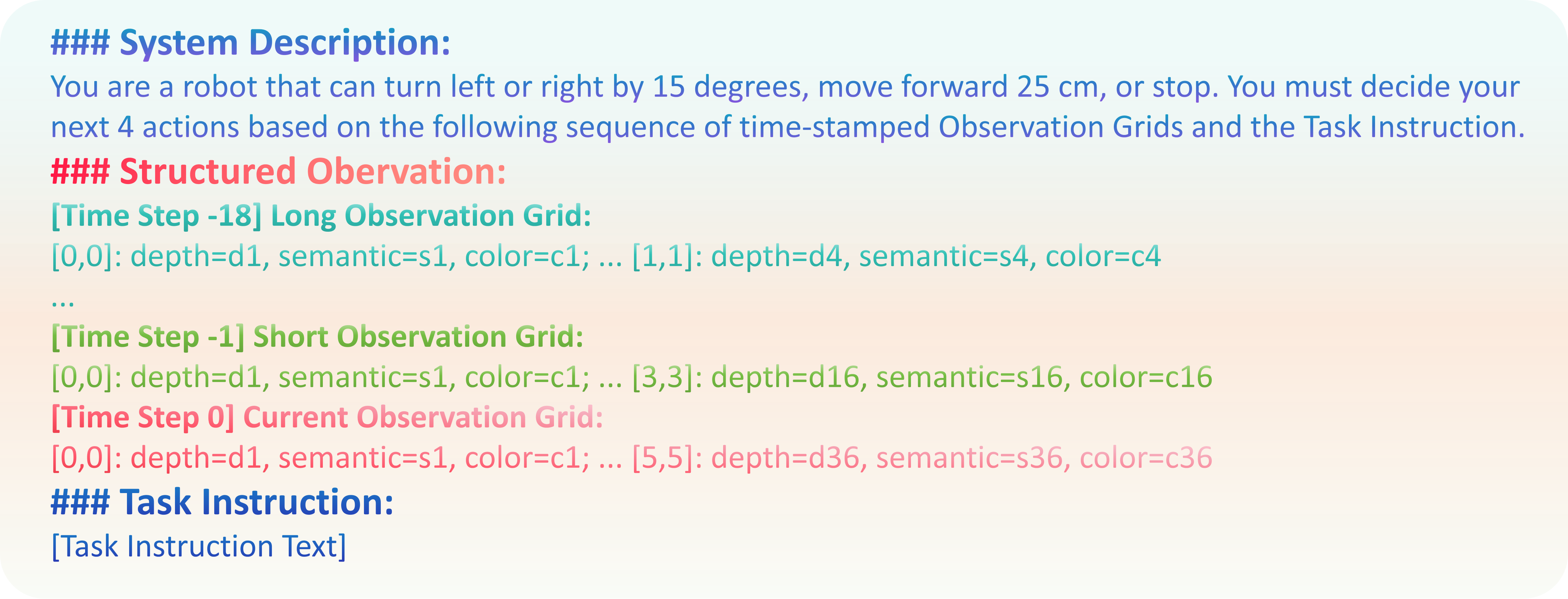}
    \caption{Structured Observation Language Prompt for LLM. The prompt integrates system description ($D_{\text{system}}$), structured observation ($O_{\text{structure}}$), and task instruction ($I_{\text{task}}$) to provide clear system definition, structured observations, and explicit prediction requirements for the language model.}
    \label{fig:prompt_example}
\end{figure}

To convert raw visual inputs into structured textual observations that can be directly understood by language models, we design a grid-based feature extraction pipeline to obtain semantic, depth, and color information from visual observations (Figure \ref{fig:prompt_example}).
% \subsubsection{Current Observation Extraction}
The semantic segmentation map of the visual observation can be obtained directly from the datasets or using pre-trained semantic segmentation model, such as SegFormer \cite{xie2021segformer}, Grounded SAM \cite{ren2024grounded}, etc. Then, we divide the RGB-D image and semantic segmentation map into $N_{\text{curr}} = 6$, $N_{\text{short}}=4$ and $N_{\text{long}}=2$ grids. For each grid cell:
\begin{itemize}
    \item \textit{Depth Information}: We calculate the average depth value of all pixels in the grid cell as the depth observation of the grid.
    \item \textit{Semantic Information}: We count the area of each semantic category in the grid cell and select the category with the largest area as the semantic label of the grid.
    \item \textit{Color Information}: We convert the RGB values of the grid cell to HSV color space, then map the HSV values to standard color names (e.g., light\_gray, yellow, blue) based on a predefined color lookup table.
\end{itemize}

The extracted information for each grid cell is formatted as a textual description: `[i,j]: depth=d, semantic=s, color=c', where $i,j$ are the grid coordinates, $d$ is the average depth, $s$ is the semantic label, and $c$ is the color name. In this work, we consider $n_{\text{short}}=2$ and $n_{\text{long}}=16$ for historical observations. 
All grid-based descriptions are concatenated with their respective time-step information at the element level, and then organized in chronological order to form the structured observation sequence $O_{\text{structure}}$. Mathematically, $O_{\text{structure}}$ is defined as a time-ordered sequence of augmented observation matrices:
\begin{align}
    O_{\text{structure}} = \left( (t_{-n_t}, \hat{O}_{-n_t}),\ \dots,\ (t_{-1}, \hat{O}_{-1}),\ (t_0, O_0) \right)
\end{align}
where $t_k$ ($k \in [-n_t, 0]$) denotes the time-step label for the $k$-th observation matrix. The parentheses $(\cdot)$ indicate an ordered sequence, and commas separate matrices in chronological order from the earliest historical frame to the current frame.
As shown in Figure \ref{fig:prompt_example}, the system description $D_{\text{system}}$, structured observation $O_{\text{structure}}$, and task instruction $I_{\text{task}}$ are combined as the structured observation language prompt for LLM. This input format is designed to be easily understood by the language model, with a clear description of the system, structured observations, and a clear prediction requirement.

\subsection{Model Architecture for Action Block Prediction}

% The LLM-based action chunk predictor can be setup using a custom module, built on the pre-trained model (e.g., Qwen3 Embedding \cite{zhang2025qwen3}, LongFormer \cite{beltagy2020longformer}, BigBird \cite{zaheer2020big}) as the backbone encoder which is suitable for long-text input and has strong reasoning capabilities. This choice leverages Qwen3’s efficient sparse attention mechanism to handle the long sequence length of structured observation-language prompts. 
A variety of pre-trained language models with strong long-text reasoning capabilities can serve as the backbone encoder for constructing the LLM-based action chunk predictor, such as Qwen \cite{yang2025qwen3}, BigBird \cite{zaheer2020big}, Longformer \cite{beltagy2020longformer}, etc. In this work, we adopt Qwen3 Embedding \cite{zhang2025qwen3} as the backbone encoder for our model, leveraging its extended context window and powerful reasoning capabilities to effectively process the long sequence length of structured observation-language prompts in SOL-Nav. 
The model architecture includes two key components:

\noindent \textbf{1) Backbone Encoder}: The pre-trained Qwen3 model is used to encode the unified structured observation language prompt into a dense semantic representation. 

\noindent \textbf{2) Multi-Step Classification Heads}: A module list of linear classification heads (one for each step in the action chunk) is attached to the encoder. Each head maps the embedding to a distribution over the 4 discrete action classes. All classification heads share the same class weight distribution for balanced training.

\subsection{Datasets and Training}

\noindent \textbf{Datasets}
We use the R2R-CE \cite{anderson2018vision} and RxR-CE \cite{ku2020room} datasets for training and evaluation with the Habitat simulator \cite{savva2019habitat}. For each trajectory in the datasets, we convert the raw RGB-D observations into structured textual observations as described in Section \ref{sec:model_setup}. The task instructions are retained as the original natural language text and augmented using LLM to get more diverse task instructions. We construct training samples by pairing the structured observations, task instructions, and ground-truth action blocks (each block contains $N_a=4$ consecutive actions from the trajectory), where 'stop' actions are added to supplement any insufficient steps in the last block of the trajectory.

\noindent \textbf{Training}
 We fine-tune the Qwen3-Embedding-0.6B model on standard VLN datasets (R2R-CE and RxR-CE) for 4-step action block prediction (\(N_a=4\)), with action classes determined from the dataset. We adopt Low-Rank Adaptation (LoRA) \cite{hu2022lora} for parameter-efficient fine-tuning and use weighted cross-entropy loss.
 The total loss is the average of weighted cross-entropy losses for each step in the 4-action block:
 \begin{align}
     \mathcal{L} = \frac{1}{4} \sum_{i=1}^4 \mathcal{L}_{CE}(y_i, \hat{y}_i; w)
 \end{align}
where \(y_i\) denotes the ground-truth action for the \(i\)-th step, \(\hat{y}_i\) is the predicted probability distribution over action classes, and \(w\) represents class weights computed via `balanced` class weighting to address potential class imbalance in the dataset. 
The model is trained using 4 RTX-4090 GPUs for about 2 weeks, with a total of 10 training epochs, and we select the model with the best performance based on validation metrics for subsequent use.

% ============================================================
% 4. Experiments
% ============================================================
\section{Experiments}
\label{sec:experiments}
\subsection{Comparison with Baselines}

\begin{table*}[t]
% ======================== 【长宽控制参数调控区】 ========================
\renewcommand\arraystretch{0.45}      % 【1. 纵向高度】：缩放行高（默认1.0，建议 0.75 ~ 0.95，数值越小越紧凑）
\setlength{\tabcolsep}{10pt}           % 【2. 列间距】：控制左右留白（默认通常为 6pt，数值越小列越紧凑）
% =======================================================================

\centering
\small
\caption{Comparison with state-of-the-art methods on VLN-CE R2R Val-Unseen split. $*$ indicates methods using the waypoint predictor from \cite{hong2022bridging}. $\dagger$ denotes methods using additional training data beyond the R2R-CE benchmarks.}
\label{tab:r2r_comparison}
\newcommand{\cmark}{\checkmark}

% 【3. 横向整体宽度】：通过修改 0.85\linewidth 控制表格占用页面总宽度的比例（如 0.8\linewidth, 0.9\linewidth, 1.0\linewidth）
\begin{minipage}{1.0\linewidth}
\centering
\resizebox{0.95\linewidth}{!}{%
\begin{tabular}{c|cccc|cccc}
\toprule
\multirow{2}{*}{Method} & \multicolumn{4}{c|}{Observation} & \multicolumn{4}{c}{R2R Val-Unseen} \\
\cmidrule(lr){2-5} \cmidrule(lr){6-9}
 & Pano. & Odo. & Depth & S.RGB & NE$\downarrow$ & OS$\uparrow$ & SR$\uparrow$ & SPL$\uparrow$ \\
\midrule
HPN+DN* \cite{krantz2021waypoint} & \cmark & \cmark & \cmark & & 6.31 & 40.0 & 36.0 & 34.0 \\
CMA* \cite{hong2022bridging} & \cmark & \cmark & \cmark & & 6.20 & 52.0 & 41.0 & 36.0 \\
Sim2Sim* \cite{krantz2022sim} & \cmark & \cmark & \cmark & & 6.07 & 52.0 & 43.0 & 36.0 \\
GridMM* \cite{wang2023gridmm} & \cmark & \cmark & \cmark & & 5.11 & 61.0 & 49.0 & 41.0 \\
ScaleVLN* \cite{wang2023scaling} & \cmark & \cmark & \cmark & & 4.80 & -- & 55.0 & 51.0 \\
ETPNav* \cite{an2024etpnav} & \cmark & \cmark & \cmark & & \textbf{4.71} & \textbf{65.0} & \textbf{57.0} & \textbf{49.0} \\
\midrule
InstructNav \cite{long2024instructnav} & \cmark & \cmark & \cmark & \cmark & 6.89 & -- & 31.0 & 24.0 \\
% AG-CMTP \cite{chen2021topological} & \cmark & \cmark & \cmark & & 7.90 & 39.2 & 23.1 & 19.1 \\
R2R-CMTP \cite{chen2021topological} & \cmark & \cmark & \cmark & & 7.90 & 38.0 & 26.4 & 22.7 \\
LAW \cite{raychaudhuri2021language} & & \cmark & \cmark & \cmark & 6.83 & 44.0 & 35.0 & 31.0 \\
CM2 \cite{georgakis2022cross} & & \cmark & \cmark & \cmark & 7.02 & 41.5 & 34.3 & 27.6 \\
WS-MGMap \cite{chen2022weakly} & & \cmark & \cmark & \cmark & 6.28 & 47.6 & 38.9 & 34.3 \\
ETPNav + FF \cite{wang2024sim} & & \cmark & \cmark & \cmark & 5.95 & 55.8 & 44.9 & 30.4 \\
% Seq2Seq \cite{krantz2020beyond} & & & \cmark & \cmark & 7.77 & 37.0 & 25.0 & 22.0 \\
% CMA \cite{krantz2020beyond} & & & \cmark & \cmark & 7.37 & 40.0 & 32.0 & 30.0 \\
MapNav \cite{zhang2025novel} & & & & \cmark & 4.93 & 53.0 & 39.7 & 37.2 \\
NaVid \cite{zhang2024navid} & & & & \cmark & 5.47 & 49.1 & 37.4 & 35.9 \\
NaVILA \cite{cheng2024navila} & & & & \cmark & 5.37 & 57.6 & 49.7 & 45.5 \\
\textbf{SOL-Nav (Ours)}  & & & \cmark & \cmark & \textbf{5.11} & \textbf{72.9} & \textbf{53.6} & \textbf{49.2} \\
\midrule
NaVILA$^\dagger$ \cite{cheng2024navila} & & & & \cmark & 5.22 & 62.5 & 54.0 & 49.0 \\
UniNaVid$^\dagger$ \cite{zhang2024uni} & & & & \cmark & 5.58 & 53.3 & 47.0 & 42.7 \\
InternVLA-N1$^\dagger$ \cite{wei2025ground} & & & \cmark & \cmark & \textbf{4.83} & \textbf{63.3} & \textbf{58.2} & \textbf{54.0} \\
\bottomrule
\end{tabular}%
}
\end{minipage}
\end{table*}

\begin{table}[t]
\renewcommand\arraystretch{0.9}
\centering
\caption{Comparison with state-of-the-art methods on the VLN-CE RxR Val-Unseen splits. $\ddagger$ indicates methods that only trained on
VLN-CE R2R. $*$ indicates methods that utilize the waypoint predictor proposed by \cite{hong2022bridging}. $\dagger$ denotes methods that employ additional training data beyond the RxR-CE benchmarks.}
\label{tab:rxr_comparison}
\newcommand{\cmark}{\checkmark}
\resizebox{\linewidth}{!}{%
\begin{tabular}{c|cccc|cccc}
\toprule
\multirow{2}{*}{Method} & \multicolumn{4}{c|}{Observation} & \multicolumn{4}{c}{RxR Val-Unseen} \\
\cmidrule(lr){2-5} \cmidrule(lr){6-9}
 & Pano. & Odo. & Depth & S.RGB & NE$\downarrow$ & OS$\uparrow$ & SR$\uparrow$ & SPL$\uparrow$  \\
\midrule
% HPN+DN* \cite{krantz2021waypoint} & \cmark & \cmark & \cmark & & - & - & - & - \\
CMA$^{\ddagger}$* \cite{hong2022bridging} &  &\cmark & \cmark &\cmark &11.7  & 10.7  &4.4 &2.5  \\
% A$^2$Nav* \cite{krantz2020beyond} & & & & \cmark & 11.8  & 5.02 &  3.5  & 3.4 \\

ETPNav$^{\ddagger}$* \cite{an2024etpnav} &  & \cmark & \cmark &\cmark & 8.79 &36.7  &25.5  &18.1  \\

CM2$^{\ddagger}$ \cite{Georgakis2022CrossmodalML} &  & \cmark & \cmark &\cmark & 8.98  &25.3  &14.4 &9.2  \\
 
LAW$^{\ddagger}$ \cite{raychaudhuri2021language} & & \cmark & \cmark & \cmark &10.87   &21.0 & 8.0  & 8.0 \\

WS-MGMap$^{\ddagger}$ \cite{Chen2022WeaklySupervisedMM} &  & \cmark & \cmark &\cmark & 9.83  & 29.8  &15.0 &12.1  \\

ETPNav + FF \cite{wang2024sim} & & \cmark & \cmark & \cmark &8.79  &36.7  &25.5   &18.1  \\
% CMA \cite{krantz2020beyond} & & & \cmark & \cmark & - & - & - & - \\
% \midrule
% NaVid Zhang et al. (2024) & & & & \cmark & - & - & - & - \\
% MapNav Zhang et al. (2025c) & & & & \cmark & - & - & - & - \\
% NaVILA Cheng et al. (2025) & & & & \cmark & - & - & - & - \\
% Seq2Seq$^{\ddagger}$ \cite{krantz2020beyond} & & & \cmark & \cmark & 11.8  & 5.02 &  3.5  & 3.4 \\

NaVid$^{\ddagger}$ \cite{zhang2024navid} & & & & \cmark &8.41  &34.5  &23.8  & 21.2 \\

% \midrule
% \midrule
NaVILA$^{\dagger}$ \cite{cheng2024navila} & & & & \cmark & 6.77 & -- &49.3  &44.0  \\
UniNaVid$^{\dagger}$ \cite{zhang2024uni} & & & & \cmark & 6.24 &55.5  & 48.7 & 40.9 \\
% InternVLA-N1 (S2)$\dagger$ & & & & \cmark & 6.41 & 49.5 & 41.8 & 62.6 \\
InternVLA-N1$^{\dagger}$  \cite{wei2025ground} & & & \cmark & \cmark & \textbf{5.91} & -- & \textbf{53.5} & \textbf{46.1}  \\

\textbf{SOL-Nav (Ours)} & & & \cmark & \cmark & 6.87 & \textbf{60.5} & 48.6 & 42.3 \\
\bottomrule
\end{tabular}%
}
\end{table}

To comprehensively evaluate the proposed method, we conduct extensive experiments on the R2R-CE \cite{anderson2018vision} and RxR-CE \cite{ku2020room} benchmarks to validate its effectiveness. Both benchmarks are built upon the VLN-CE setting \cite{Krantz2020BeyondTN} and are implemented using the Habitat simulator. They simulate realistic indoor navigation scenarios in MP3D environments, where agents are required to follow natural language instructions under continuous control.
R2R-CE provides English-only instructions with relatively short navigation paths, whereas RxR-CE is a large-scale multilingual benchmark featuring longer and more diverse trajectories.
To assess the generalization capability of our approach, all experiments are conducted on the validation unseen (val-unseen) splits of both benchmarks. For fair comparison, we follow prior work and adopt the standard VLN evaluation metrics:
\begin{itemize}
\item Navigation Error (NE):the final distance between the agent’s stopping position and the goal location;
\item Success Rate (SR): the percentage of episodes in which the agent stops within 3 meters of the goal;
\item Oracle Success Rate (OS): counts an episode as successful if any point along the trajectory satisfies the success criterion;
\item Success weighted by Path Length (SPL): a success metric that penalizes unnecessarily long trajectories.

\end{itemize}

Together, these metrics provide a comprehensive evaluation of instruction-following performance in terms of both effectiveness and efficiency.

\textcolor{black}{We compare our SOL-Nav model with a diverse set of baseline methods. These baselines include single-modal approaches such as NaVid \cite{zhang2024navid}, UniNaVid \cite{zhang2024uni}, NaVILA \cite{cheng2024navila}, and MapNav \cite{zhang2025novel}, as well as multimodal methods like CMA \cite{hong2022bridging}, ETPNav \cite{an2024etpnav}, and InternVLA-N1 \cite{wei2025ground}, among others. The experimental results are presented in Table \ref{tab:r2r_comparison} and Table \ref{tab:rxr_comparison}, where Table \ref{tab:r2r_comparison} reports the performance on the R2R Val-Unseen split, and Table \ref{tab:rxr_comparison} shows the results on the RxR Val-Unseen split.}

As shown in Table \ref{tab:r2r_comparison}, our method achieves highly competitive performance. Specifically, when compared with approaches that do not employ the waypoint predictor from \cite{hong2022bridging} and do not use additional training data beyond the R2R-CE benchmarks (the second column of Table \ref{tab:r2r_comparison}), our SOL-Nav attains the best overall results. In particular, it achieves state-of-the-art performance on the four metrics NE, OS, SR, and SPL, reaching 5.11, 72.9, 53.6, and 49.2, respectively.
When compared with the methods in the first column of Table \ref{tab:r2r_comparison} that utilize the waypoint predictor from \cite{hong2022bridging}, SOL-Nav exhibits only marginal differences relative to ETPNav on OS, SR, and SPL. Notably, our SOL-Nav outperforms ETPNav by more than 12\% on the OS metric, while not relying on the waypoint predictor introduced in \cite{hong2022bridging}.
Compared with the methods listed in the third column, SOL-Nav performs slightly worse than InternVLA-N1 on OS, SR, and SPL. However, on the OS metric, our approach surpasses InternVLA-N1 by more than 15\%. It is worth emphasizing that all methods in the third column are trained with additional data, whereas SOL-Nav is trained solely on the R2R-CE benchmarks. Moreover, our model contains only 0.6B parameters, while the other methods in the third column have model sizes of at least 7B. Our model is more than 10 times smaller in scale, which is of significant practical importance for real-world deployment.

% The experimental results on the VLN-CE RxR Val-Unseen split exhibit trends similar to those described above. The quantitative results are presented in Table \ref{tab:rxr_comparison}.
% Specifically, as shown in Table \ref{tab:rxr_comparison}, our SOL-Nav achieves 6.95, 60.5, 48.6, and 42.3 on NE, OS, SP, and SPL, respectively. Compared with the methods listed in the first and second columns, our approach delivers the best overall performance, even though the methods in the first column employ a waypoint predictor. When compared with the methods in the third column, our results are slightly lower than those of NaVILA \cite{cheng2024navila} and InternVLA-N1 \cite{wei2025ground} on NE, SR, and SPL. However, relative to UniNaVid \cite{zhang2024uni}, our method outperforms it on both OS and SPL, achieving an advantage of nearly 10\% on the OS metric.

Experimental results on the VLN-CE RxR Val-Unseen split exhibit trends consistent with those described above, with detailed quantitative results presented in Table \ref{tab:rxr_comparison}. As shown in Table \ref{tab:rxr_comparison}, our proposed SOL-Nav method achieves 6.95, 60.5, 48.6, and 42.3 on Navigation Error (NE), Objective Success (OS), Success weighted by Path Length (SP), and Success weighted by Path Length (SPL), respectively. Although our results are slightly lower than those of NaVILA and InternVLA-N1 on NE, SR, and SPL metrics, our method outperforms UniNaVid on both OS and SPL, achieving a nearly 10\% performance advantage on the OS metric. It is worth noting that, unlike UniNaVid, NaVILA, and InternVLA-N1, our method achieves the competitive experimental results without being trained on additional datasets.

Considering the overall experimental results on the R2R Val-Unseen and RxR Val-Unseen splits, our method achieves highly competitive performance without employing a waypoint predictor or utilizing additional training data. In terms of overall metrics, it is only slightly inferior to the latest models trained with extra data. However, in particular, our model is more than 50 times smaller than these counterparts, which is of substantial practical significance for real-world robotic deployment.
In addition, our approach significantly outperforms all state-of-the-art models on the OS metric, further demonstrating its strong potential. Future work will focus on improving the stop prediction capability to further enhance overall performance.

\subsection{Ablation Study}
We conduct ablation studies on the R2R-CE \textit{val-unseen} split to evaluate key components of SOL-Nav (Table~\ref{tab:ablation}). The full model achieves the best overall performance ($\text{NE}=5.11$, $\text{SR}=53.6\%$, $\text{SPL}=49.2\%$). 

Ablating \textbf{semantic labels} causes the most severe drop ($\text{SR}\rightarrow20.3\%$), confirming semantics as the primary alignment bridge. Removing \textbf{depth} ($\text{SR}\rightarrow21.6\%$) significantly impairs 3D spatial reasoning, while omitting \textbf{history} ($\text{SR}\rightarrow26.5\%$) leads to short-sighted decisions. Reducing \textbf{grid resolution} to $4\times4$ ($\text{SR}\rightarrow34.5\%$) causes loss of fine-grained spatial detail. Finally, removing \textbf{color} ($\text{SR}\rightarrow48.7\%$) shows a moderate decline, indicating its auxiliary grounding value. Overall, these results confirm that all components contribute synergistically to optimal performance.

\begin{table}[t]
    \centering
    \small  % 缩小字号以适应新增列，避免溢出
    \caption{Ablation study results on the R2R-CE val-unseen split, evaluating the impact of five core components of SOL-Nav: grid resolution (6×6 vs. 4×4), historical observation, depth, semantic segmentation, and color information. The full model (6×6 resolution, with all five inputs) serves as the baseline for all ablation variants.}
    \label{tab:ablation}
    \newcommand{\cmark}{\checkmark}
    \resizebox{\linewidth}{!}{%
    \begin{tabular}{l|ccccc|cccc} % 调整为 1 + 2 + 3 + 4 列，视觉上分组更清晰
        \toprule
        \multirow{2}{*}{Ablation Term} & \multicolumn{5}{c|}{Observation Config} & \multicolumn{4}{c}{Metrics} \\
        \cline{2-10} 
                         & Resolution & History & Depth & Semantic & Color & NE$\downarrow$ & OS$\uparrow$ & SR$\uparrow$ & SPL$\uparrow$ \\
        \midrule
        Lower Res.       & $4 \times 4$ & $\cmark$ & $\cmark$ & $\cmark$ & $\cmark$ & 6.84  & 43.4 & 34.5 & 29.8 \\
        No His.          & $6 \times 6$ &          & $\cmark$ & $\cmark$ & $\cmark$ & 7.81  & 39.4 & 26.5 & 21.9 \\
        No Depth         & $6 \times 6$ & $\cmark$ &          & $\cmark$ & $\cmark$ & 7.98  & 31.2 & 21.6 & 17.8 \\
        No Semantic      & $6 \times 6$ & $\cmark$ & $\cmark$ &          & $\cmark$ & 8.30  & 28.1 & 20.3 &15.2 \\ 
        No Color         & $6 \times 6$ & $\cmark$ & $\cmark$ & $\cmark$ &   & 7.45  &63.8  & 48.7 & 41.5 \\ 
        All Info.        & $6 \times 6$ & $\cmark$ & $\cmark$ & $\cmark$ & $\cmark$ & \textbf{5.11} & \textbf{72.9} & \textbf{53.6} & \textbf{49.2} \\
        \bottomrule
    \end{tabular}%
    }
\end{table}

\begin{table}[htbp]
  \centering
  \small
  \caption{Computational cost comparison results.}
  \label{tab:infer_time_compare}
  \renewcommand{\arraystretch}{0.65}
  \setlength{\tabcolsep}{4pt}
  \begin{tabular}{lcc}
    \toprule
    Method & Total Latency (s) $\downarrow$ & GPU Memory (GB) $\downarrow$ \\
    \midrule
    NaVILA       & 0.595          & 18.5 \\
    InternVLA-N1 & 0.730          & 20.0 \\
    \textbf{Ours} & \textbf{0.091} & \textbf{2.2} \\
    \bottomrule
  \end{tabular}
\end{table}

% \begin{table}[htbp]
%   \centering
%   \caption{Computational cost comparison results.}
%   \label{tab:infer_time_compare}
%   % 0.6\linewidth 表示缩放到当前页面文本宽度的 60%
%   \resizebox{1.0\linewidth}{!}{% 
%     \begin{tabular}{lcc}
%       \toprule
%       & \multicolumn{2}{c}{Computational Cost} \\
%       \cmidrule(lr){2-3}
%       & Total Latency (s) $\downarrow$ & GPU Memory (GB) $\downarrow$ \\
%       \midrule
%       NaVILA  & 0.595          & 18.5 \\
%       % \rowcolor{lightgreen}
%       InternVLA-N1 & 0.730 & 20.0 \\
%       Ours & \textbf{0.091} & \textbf{2.2} \\
%       \bottomrule
%     \end{tabular}%
%   }
% \end{table}

\begin{table}[htbp]
    \centering
    \small  % 缩小字号以适应半栏
    \caption{Inference latency (s) details on different devices.}
    \label{inference_time}
    \setlength{\tabcolsep}{6pt}
    \begin{tabular}{lcccc}
        \toprule
        Device & Segmentation & SOP & Qwen3-E & Total \\
        \midrule
        Jetson Orin & 0.032 & 0.220 & 0.533 & 0.785 \\
        RTX 4090    & 0.003 & 0.023 & 0.065 & 0.091 \\
        \bottomrule
    \end{tabular}
    \label{tab:detail_latency}
\end{table}

\subsection{Inference Latency Analysis}

We compare the computational overhead with SOTA methods in Table \ref{tab:infer_time_compare}. Our method exhibits far lower inference latency and memory consumption, indicating its strong potential for real-world deployment.
The time consumption of our method primarily stems from three components: semantic segmentation, the Structured Observation Preprocessor (SOP), and Qwen3-Embedding. A detailed breakdown of the runtime for each component is presented in Table \ref{tab:detail_latency}.

\subsection{Real-World Deployments}

\begin{figure}
    \centering
    \includegraphics[width=1.0\linewidth]{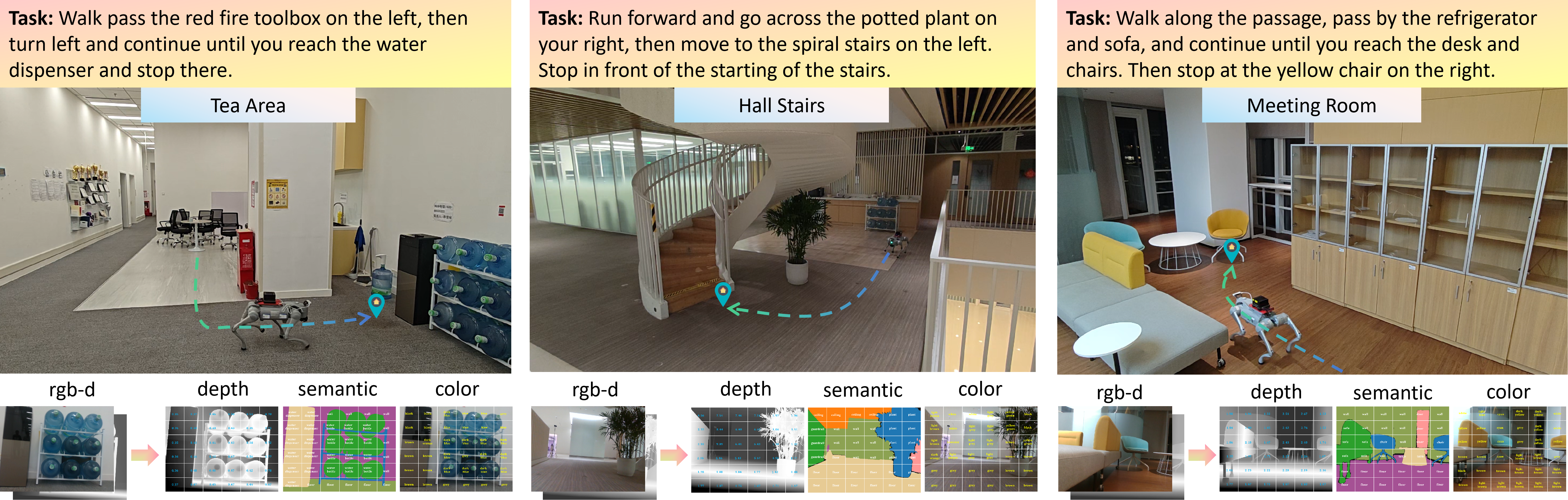}
    \caption{Real-world Deployments. We conduct real-world navigation experiments in three distinct scenarios with varying environmental characteristics (Tea Area, Hall Stairs, Meeting Room), to comprehensively evaluate the robustness and generalization of SOL-Nav.}
    \label{fig:real_experiments}
\end{figure}

To validate the practical applicability of SOL-Nav, we carried out real-world deployment experiments using the Unitree Go2 robot platform, with NVIDIA Jetson Orin as the edge computing unit and Intel RealSense D435i (640×480 resolution) for the acquisition of RGB-D images. 
We use SegFormer \cite{xie2021segformer} for high-quality semantic segmentation. 
% We further fine-tune the pre-trained SOL-Nav model using 50 real-world navigation samples mixed with partial R2R data. 
Experiments are conducted in three scenarios (Tea Area, Hall Stairs, Meeting Room) as shown in Figure \ref{fig:real_experiments}. The inference latency is about 0.8s on Jetson Orin, meeting the real-time requirement for robotic navigation. The results confirm SOL-Nav’s real-time performance and robustness against real-world environmental variations.

% ============================================================
% 5. Conclusion
% ============================================================
\section{Conclusion}
\label{sec:conclusion}

This paper proposes SOL-Nav, a novel VLN framework that converts egocentric RGB-D observations into grid-structured textual descriptions (semantic, color, and depth), enabling pure language model-based decision-making. By decoupling high-level semantics from raw visual features, SOL-Nav fully leverages pre-trained language models without requiring visual encoder training from scratch or complex multimodal fusion. Experiments on R2R-CE and RxR-CE show that SOL-Nav achieves competitive or state-of-the-art performance—outperforming existing models on the OS metric—while using a significantly smaller model (0.6B parameters) and less training data.

However, relying solely on discrete semantic, color, and depth features leads to the loss of fine-grained visual details (e.g., shape, texture), which may affect precise navigation in complex scenes. Future work will incorporate richer visual cues, explore adaptive grid resolutions, and optimize model efficiency for edge device deployment.

Overall, SOL-Nav establishes an efficient and generalizable paradigm for vision-language navigation, offering strong potential for broader embodied AI applications such as robotic manipulation.

\bibliography{aaai2027}

@String(CVPR  = {IEEE Conf. Comput. Vis. Pattern Recog.})

@String(NeurIPS = {Adv. Neural Inform. Process. Syst.})

@String(ICLR  = {Int. Conf. Learn. Represent.})

@String(AAAI  = {AAAI})

@String(CVPR  = {CVPR})

@String(NeurIPS = {NeurIPS})

@String(ICLR  = {ICLR})

@article{gao2025octonav,
  title={OctoNav: Towards Generalist Embodied Navigation},
  author={Gao, Chen and Jin, Liankai and Peng, Xingyu and Zhang, Jiazhao and Deng, Yue and Li, Annan and Wang, He and Liu, Si},
  journal={arXiv preprint arXiv:2506.09839},
  year={2025}
}

@article{wei2025streamvln,
  title={Streamvln: Streaming vision-and-language navigation via slowfast context modeling},
  author={Wei, Meng and Wan, Chenyang and Yu, Xiqian and Wang, Tai and Yang, Yuqiang and Mao, Xiaohan and Zhu, Chenming and Cai, Wenzhe and Wang, Hanqing and Chen, Yilun and others},
  journal={arXiv preprint arXiv:2507.05240},
  year={2025}
}

@article{zhang2024navid,
  title={Navid: Video-based vlm plans the next step for vision-and-language navigation},
  author={Zhang, Jiazhao and Wang, Kunyu and Xu, Rongtao and Zhou, Gengze and Hong, Yicong and Fang, Xiaomeng and Wu, Qi and Zhang, Zhizheng and Wang, He},
  journal={arXiv preprint arXiv:2402.15852},
  year={2024}
}

@article{peng2025lovon,
  title={Lovon: Legged open-vocabulary object navigator},
  author={Peng, Daojie and Cao, Jiahang and Zhang, Qiang and Ma, Jun},
  journal={arXiv preprint arXiv:2507.06747},
  year={2025}
}

@inproceedings{zhu2017target,
  title={Target-driven visual navigation in indoor scenes using deep reinforcement learning},
  author={Zhu, Yuke and Mottaghi, Roozbeh and Kolve, Eric and Lim, Joseph J and Gupta, Abhinav and Fei-Fei, Li and Farhadi, Ali},
  booktitle={2017 IEEE international conference on robotics and automation (ICRA)},
  pages={3357--3364},
  year={2017},
  organization={IEEE}
}

@article{wang2025monodream,
  title={Monodream: Monocular vision-language navigation with panoramic dreaming},
  author={Wang, Shuo and Wang, Yongcai and Fan, Zhaoxin and Wang, Yucheng and Chen, Maiyue and Wang, Kaihui and Su, Zhizhong and Li, Wanting and Cai, Xudong and Jin, Yeying and others},
  journal={arXiv preprint arXiv:2508.02549},
  year={2025}
}

@article{chang2017matterport3d,
  title={Matterport3d: Learning from rgb-d data in indoor environments},
  author={Chang, Angel and Dai, Angela and Funkhouser, Thomas and Halber, Maciej and Niessner, Matthias and Savva, Manolis and Song, Shuran and Zeng, Andy and Zhang, Yinda},
  journal={arXiv preprint arXiv:1709.06158},
  year={2017}
}

@article{mavrogiannis2023core,
  title={Core challenges of social robot navigation: A survey},
  author={Mavrogiannis, Christoforos and Baldini, Francesca and Wang, Allan and Zhao, Dapeng and Trautman, Pete and Steinfeld, Aaron and Oh, Jean},
  journal={ACM Transactions on Human-Robot Interaction},
  volume={12},
  number={3},
  pages={1--39},
  year={2023},
  publisher={ACM New York, NY}
}

@inproceedings{zheng2024towards,
  title={Towards learning a generalist model for embodied navigation},
  author={Zheng, Duo and Huang, Shijia and Zhao, Lin and Zhong, Yiwu and Wang, Liwei},
  booktitle={Proceedings of the IEEE/CVF Conference on Computer Vision and Pattern Recognition},
  pages={13624--13634},
  year={2024}
}

@article{duan2022survey,
  title={A survey of embodied ai: From simulators to research tasks},
  author={Duan, Jiafei and Yu, Samson and Tan, Hui Li and Zhu, Hongyuan and Tan, Cheston},
  journal={IEEE Transactions on Emerging Topics in Computational Intelligence},
  volume={6},
  number={2},
  pages={230--244},
  year={2022},
  publisher={IEEE}
}

@article{ku2020room,
  title={Room-across-room: Multilingual vision-and-language navigation with dense spatiotemporal grounding},
  author={Ku, Alexander and Anderson, Peter and Patel, Roma and Ie, Eugene and Baldridge, Jason},
  journal={arXiv preprint arXiv:2010.07954},
  year={2020}
}

@article{liu2025aligning,
  title={Aligning cyber space with physical world: A comprehensive survey on embodied ai},
  author={Liu, Yang and Chen, Weixing and Bai, Yongjie and Liang, Xiaodan and Li, Guanbin and Gao, Wen and Lin, Liang},
  journal={IEEE/ASME Transactions on Mechatronics},
  year={2025},
  publisher={IEEE}
}

@article{zhang2024uni,
  title={Uni-navid: A video-based vision-language-action model for unifying embodied navigation tasks},
  author={Zhang, Jiazhao and Wang, Kunyu and Wang, Shaoan and Li, Minghan and Liu, Haoran and Wei, Songlin and Wang, Zhongyuan and Zhang, Zhizheng and Wang, He},
  journal={arXiv preprint arXiv:2412.06224},
  year={2024}
}

@inproceedings{anderson2018vision,
  title={Vision-and-language navigation: Interpreting visually-grounded navigation instructions in real environments},
  author={Anderson, Peter and Wu, Qi and Teney, Damien and Bruce, Jake and Johnson, Mark and S{\"u}nderhauf, Niko and Reid, Ian and Gould, Stephen and Van Den Hengel, Anton},
  booktitle={Proceedings of the IEEE conference on computer vision and pattern recognition},
  pages={3674--3683},
  year={2018}
}

@article{luo2025navimaster,
  title={Navimaster: Learning a unified policy for gui and embodied navigation tasks},
  author={Luo, Zhihao and Yan, Wentao and Gong, Jingyu and Wang, Min and Zhang, Zhizhong and Wang, Xuhong and Xie, Yuan and Tan, Xin},
  journal={arXiv preprint arXiv:2508.02046},
  year={2025}
}

@article{yuan2024robopoint,
  title={Robopoint: A vision-language model for spatial affordance prediction for robotics},
  author={Yuan, Wentao and Duan, Jiafei and Blukis, Valts and Pumacay, Wilbert and Krishna, Ranjay and Murali, Adithyavairavan and Mousavian, Arsalan and Fox, Dieter},
  journal={arXiv preprint arXiv:2406.10721},
  year={2024}
}

@article{ramakrishnan2021habitat,
  title={Habitat-matterport 3d dataset (hm3d): 1000 large-scale 3d environments for embodied ai},
  author={Ramakrishnan, Santhosh K and Gokaslan, Aaron and Wijmans, Erik and Maksymets, Oleksandr and Clegg, Alex and Turner, John and Undersander, Eric and Galuba, Wojciech and Westbury, Andrew and Chang, Angel X and others},
  journal={arXiv preprint arXiv:2109.08238},
  year={2021}
}

@article{zhu2021deep,
  title={Deep learning for embodied vision navigation: A survey},
  author={Zhu, Fengda and Zhu, Yi and Lee, Vincent and Liang, Xiaodan and Chang, Xiaojun},
  journal={arXiv preprint arXiv:2108.04097},
  year={2021}
}

@article{bu2025univla,
  title={Univla: Learning to act anywhere with task-centric latent actions},
  author={Bu, Qingwen and Yang, Yanting and Cai, Jisong and Gao, Shenyuan and Ren, Guanghui and Yao, Maoqing and Luo, Ping and Li, Hongyang},
  journal={arXiv preprint arXiv:2505.06111},
  year={2025}
}

@article{wang2025trackvla,
  title={Trackvla: Embodied visual tracking in the wild},
  author={Wang, Shaoan and Zhang, Jiazhao and Li, Minghan and Liu, Jiahang and Li, Anqi and Wu, Kui and Zhong, Fangwei and Yu, Junzhi and Zhang, Zhizheng and Wang, He},
  journal={arXiv preprint arXiv:2505.23189},
  year={2025}
}

@inproceedings{wijmans2019embodied,
  title={Embodied question answering in photorealistic environments with point cloud perception},
  author={Wijmans, Erik and Datta, Samyak and Maksymets, Oleksandr and Das, Abhishek and Gkioxari, Georgia and Lee, Stefan and Essa, Irfan and Parikh, Devi and Batra, Dhruv},
  booktitle={Proceedings of the IEEE/CVF Conference on Computer Vision and Pattern Recognition},
  pages={6659--6668},
  year={2019}
}

@inproceedings{yu2019multi,
  title={Multi-target embodied question answering},
  author={Yu, Licheng and Chen, Xinlei and Gkioxari, Georgia and Bansal, Mohit and Berg, Tamara L and Batra, Dhruv},
  booktitle={Proceedings of the IEEE/CVF Conference on Computer Vision and Pattern Recognition},
  pages={6309--6318},
  year={2019}
}

@article{tan2023knowledge,
  title={Knowledge-based embodied question answering},
  author={Tan, Sinan and Ge, Mengmeng and Guo, Di and Liu, Huaping and Sun, Fuchun},
  journal={IEEE Transactions on Pattern Analysis and Machine Intelligence},
  volume={45},
  number={10},
  pages={11948--11960},
  year={2023},
  publisher={IEEE}
}

@inproceedings{islam2023eqa,
  title={Eqa-mx: Embodied question answering using multimodal expression},
  author={Islam, Md Mofijul and Gladstone, Alexi and Islam, Riashat and Iqbal, Tariq},
  booktitle={The Twelfth International Conference on Learning Representations},
  year={2023}
}

@inproceedings{krantz2021waypoint,
  title={Waypoint models for instruction-guided navigation in continuous environments},
  author={Krantz, Jacob and Gokaslan, Aaron and Batra, Dhruv and Lee, Stefan and Maksymets, Oleksandr},
  booktitle={Proceedings of the IEEE/CVF International Conference on Computer Vision},
  pages={15162--15171},
  year={2021}
}

@inproceedings{hong2022bridging,
  title={Bridging the gap between learning in discrete and continuous environments for vision-and-language navigation},
  author={Hong, Yicong and Wang, Zun and Wu, Qi and Gould, Stephen},
  booktitle={Proceedings of the IEEE/CVF conference on computer vision and pattern recognition},
  pages={15439--15449},
  year={2022}
}

@inproceedings{krantz2022sim,
  title={Sim-2-sim transfer for vision-and-language navigation in continuous environments},
  author={Krantz, Jacob and Lee, Stefan},
  booktitle={European conference on computer vision},
  pages={588--603},
  year={2022},
  organization={Springer}
}

@inproceedings{wang2023gridmm,
  title={Gridmm: Grid memory map for vision-and-language navigation},
  author={Wang, Zihan and Li, Xiangyang and Yang, Jiahao and Liu, Yeqi and Jiang, Shuqiang},
  booktitle={Proceedings of the IEEE/CVF International conference on computer vision},
  pages={15625--15636},
  year={2023}
}

@article{an2024etpnav,
  title={Etpnav: Evolving topological planning for vision-language navigation in continuous environments},
  author={An, Dong and Wang, Hanqing and Wang, Wenguan and Wang, Zun and Huang, Yan and He, Keji and Wang, Liang},
  journal={IEEE Transactions on Pattern Analysis and Machine Intelligence},
  year={2024},
  publisher={IEEE}
}

@inproceedings{wang2023scaling,
  title={Scaling data generation in vision-and-language navigation},
  author={Wang, Zun and Li, Jialu and Hong, Yicong and Wang, Yi and Wu, Qi and Bansal, Mohit and Gould, Stephen and Tan, Hao and Qiao, Yu},
  booktitle={Proceedings of the IEEE/CVF international conference on computer vision},
  pages={12009--12020},
  year={2023}
}

@article{long2024instructnav,
  title={Instructnav: Zero-shot system for generic instruction navigation in unexplored environment},
  author={Long, Yuxing and Cai, Wenzhe and Wang, Hongcheng and Zhan, Guanqi and Dong, Hao},
  journal={arXiv preprint arXiv:2406.04882},
  year={2024}
}

@inproceedings{chen2021topological,
  title={Topological planning with transformers for vision-and-language navigation},
  author={Chen, Kevin and Chen, Junshen K and Chuang, Jo and V{\'a}zquez, Marynel and Savarese, Silvio},
  booktitle={Proceedings of the IEEE/CVF Conference on Computer Vision and Pattern Recognition},
  pages={11276--11286},
  year={2021}
}

@inproceedings{raychaudhuri2021language,
  title={Language-aligned waypoint (law) supervision for vision-and-language navigation in continuous environments},
  author={Raychaudhuri, Sonia and Wani, Saim and Patel, Shivansh and Jain, Unnat and Chang, Angel},
  booktitle={Proceedings of the 2021 conference on empirical methods in natural language processing},
  pages={4018--4028},
  year={2021}
}

@inproceedings{georgakis2022cross,
  title={Cross-modal map learning for vision and language navigation},
  author={Georgakis, Georgios and Schmeckpeper, Karl and Wanchoo, Karan and Dan, Soham and Miltsakaki, Eleni and Roth, Dan and Daniilidis, Kostas},
  booktitle={Proceedings of the IEEE/CVF conference on computer vision and pattern recognition},
  pages={15460--15470},
  year={2022}
}

@article{chen2022weakly,
  title={Weakly-supervised multi-granularity map learning for vision-and-language navigation},
  author={Chen, Peihao and Ji, Dongyu and Lin, Kunyang and Zeng, Runhao and Li, Thomas and Tan, Mingkui and Gan, Chuang},
  journal={Advances in Neural Information Processing Systems},
  volume={35},
  pages={38149--38161},
  year={2022}
}

@article{wang2024sim,
  title={Sim-to-real transfer via 3d feature fields for vision-and-language navigation},
  author={Wang, Zihan and Li, Xiangyang and Yang, Jiahao and Liu, Yeqi and Jiang, Shuqiang},
  journal={arXiv preprint arXiv:2406.09798},
  year={2024}
}

@article{zhang2025novel,
  title={A novel memory representation via annotated semantic maps for vlm-based vision-and-language navigation},
  author={Zhang, L and Hao, X and Xu, Q and Zhang, Q and Zhang, X and Wang, P and Zhang, J and Wang, Z and Zhang, S and Xu, R MapNav},
  journal={arXiv preprint arXiv:2502.13451},
  year={2025}
}

@article{cheng2024navila,
  title={Navila: Legged robot vision-language-action model for navigation},
  author={Cheng, An-Chieh and Ji, Yandong and Yang, Zhaojing and Gongye, Zaitian and Zou, Xueyan and Kautz, Jan and B{\i}y{\i}k, Erdem and Yin, Hongxu and Liu, Sifei and Wang, Xiaolong},
  journal={arXiv preprint arXiv:2412.04453},
  year={2024}
}

@article{wei2025ground,
  title={Ground slow, move fast: A dual-system foundation model for generalizable vision-and-language navigation},
  author={Wei, Meng and Wan, Chenyang and Peng, Jiaqi and Yu, Xiqian and Yang, Yuqiang and Feng, Delin and Cai, Wenzhe and Zhu, Chenming and Wang, Tai and Pang, Jiangmiao and others},
  journal={arXiv preprint arXiv:2512.08186},
  year={2025}
}

@article{lin2025navcot,
  title={Navcot: Boosting llm-based vision-and-language navigation via learning disentangled reasoning},
  author={Lin, Bingqian and Nie, Yunshuang and Wei, Ziming and Chen, Jiaqi and Ma, Shikui and Han, Jianhua and Xu, Hang and Chang, Xiaojun and Liang, Xiaodan},
  journal={IEEE Transactions on Pattern Analysis and Machine Intelligence},
  year={2025},
  publisher={IEEE}
}

@article{wu2024vision,
  title={Vision-language navigation: a survey and taxonomy},
  author={Wu, Wansen and Chang, Tao and Li, Xinmeng and Yin, Quanjun and Hu, Yue},
  journal={Neural Computing and Applications},
  volume={36},
  number={7},
  pages={3291--3316},
  year={2024},
  publisher={Springer}
}

@article{goetting2024end,
  title={End-to-end navigation with vision language models: Transforming spatial reasoning into question-answering},
  author={Goetting, Dylan and Singh, Himanshu Gaurav and Loquercio, Antonio},
  journal={arXiv preprint arXiv:2411.05755},
  year={2024}
}

@inproceedings{qi2020reverie,
  title={Reverie: Remote embodied visual referring expression in real indoor environments},
  author={Qi, Yuankai and Wu, Qi and Anderson, Peter and Wang, Xin and Wang, William Yang and Shen, Chunhua and Hengel, Anton van den},
  booktitle={Proceedings of the IEEE/CVF conference on computer vision and pattern recognition},
  pages={9982--9991},
  year={2020}
}

@inproceedings{savva2019habitat,
  title={Habitat: A platform for embodied ai research},
  author={Savva, Manolis and Kadian, Abhishek and Maksymets, Oleksandr and Zhao, Yili and Wijmans, Erik and Jain, Bhavana and Straub, Julian and Liu, Jia and Koltun, Vladlen and Malik, Jitendra and others},
  booktitle={Proceedings of the IEEE/CVF international conference on computer vision},
  pages={9339--9347},
  year={2019}
}

@inproceedings{chen2024mapgpt,
  title={Mapgpt: Map-guided prompting with adaptive path planning for vision-and-language navigation},
  author={Chen, Jiaqi and Lin, Bingqian and Xu, Ran and Chai, Zhenhua and Liang, Xiaodan and Wong, Kwan-Yee},
  booktitle={Proceedings of the 62nd Annual Meeting of the Association for Computational Linguistics (Volume 1: Long Papers)},
  pages={9796--9810},
  year={2024}
}

@inproceedings{qiao2025open,
  title={Open-nav: Exploring zero-shot vision-and-language navigation in continuous environment with open-source llms},
  author={Qiao, Yanyuan and Lyu, Wenqi and Wang, Hui and Wang, Zixu and Li, Zerui and Zhang, Yuan and Tan, Mingkui and Wu, Qi},
  booktitle={2025 IEEE International Conference on Robotics and Automation (ICRA)},
  pages={6710--6717},
  year={2025},
  organization={IEEE}
}

@article{zhang2025nava,
  title={NavA$^3$: Understanding Any Instruction, Navigating Anywhere, Finding Anything},
  author={Zhang, Lingfeng and Hao, Xiaoshuai and Tang, Yingbo and Fu, Haoxiang and Zheng, Xinyu and Wang, Pengwei and Wang, Zhongyuan and Ding, Wenbo and Zhang, Shanghang},
  journal={arXiv preprint arXiv:2508.04598},
  year={2025}
}

@inproceedings{zhou2024navgpt,
  title={Navgpt: Explicit reasoning in vision-and-language navigation with large language models},
  author={Zhou, Gengze and Hong, Yicong and Wu, Qi},
  booktitle={Proceedings of the AAAI Conference on Artificial Intelligence},
  volume={38},
  number={7},
  pages={7641--7649},
  year={2024}
}

@article{zaheer2020big,
  title={Big bird: Transformers for longer sequences},
  author={Zaheer, Manzil and Guruganesh, Guru and Dubey, Kumar Avinava and Ainslie, Joshua and Alberti, Chris and Ontanon, Santiago and Pham, Philip and Ravula, Anirudh and Wang, Qifan and Yang, Li and others},
  journal={Advances in neural information processing systems},
  volume={33},
  pages={17283--17297},
  year={2020}
}

@article{Georgakis2022CrossmodalML,
  title={Cross-modal Map Learning for Vision and Language Navigation},
  author={Georgios Georgakis and Karl Schmeckpeper and Karan Wanchoo and Soham Dan and Eleni Miltsakaki and Dan Roth and Kostas Daniilidis},
  journal={2022 IEEE/CVF Conference on Computer Vision and Pattern Recognition (CVPR)},
  year={2022},
  pages={15439-15449}
}

@article{Chen2022WeaklySupervisedMM,
  title={Weakly-Supervised Multi-Granularity Map Learning for Vision-and-Language Navigation},
  author={Peihao Chen and Dongyu Ji and Kun-Li Channing Lin and Runhao Zeng and Thomas H. Li and Mingkui Tan and Chuang Gan},
  journal={ArXiv},
  year={2022},
  volume={abs/2210.07506}
}

@article{li2025yolosam,
  title={YOLO-SAM: an end-to-end framework for efficient real time object detection and segmentation},
  author={Li, X. and Pu, X. and Ling, W. and others},
  journal={Scientific Reports},
  volume={15},
  number={1},
  pages={40854},
  year={2025},
  month={nov},
  publisher={Nature Publishing Group},
  doi={10.1038/s41598-025-24576-6},
  url={https://doi.org/10.1038/s41598-025-24576-6}
}

@article{ren2024grounded,
  title={Grounded sam: Assembling open-world models for diverse visual tasks},
  author={Ren, Tianhe and Liu, Shilong and Zeng, Ailing and Lin, Jing and Li, Kunchang and Cao, He and Chen, Jiayu and Huang, Xinyu and Chen, Yukang and Yan, Feng and others},
  journal={arXiv preprint arXiv:2401.14159},
  year={2024}
}

@inproceedings{Krantz2020BeyondTN,
  title={Beyond the Nav-Graph: Vision-and-Language Navigation in Continuous Environments},
  author={Jacob Krantz and Erik Wijmans and Arjun Majumdar and Dhruv Batra and Stefan Lee},
  booktitle={European Conference on Computer Vision},
  year={2020},
  url={https://api.semanticscholar.org/CorpusID:214802389}
}

@inproceedings{xie2021segformer,
  title={SegFormer: Simple and Efficient Design for Semantic Segmentation with Transformers},
  author={Xie, Enze and Wang, Wenhai and Yu, Zhiding and Anandkumar, Anima and Alvarez, Jose M and Luo, Ping},
  booktitle={Neural Information Processing Systems (NeurIPS)},
  year={2021}
}

@article{zhang2025qwen3,
  title={Qwen3 embedding: Advancing text embedding and reranking through foundation models},
  author={Zhang, Yanzhao and Li, Mingxin and Long, Dingkun and Zhang, Xin and Lin, Huan and Yang, Baosong and Xie, Pengjun and Yang, An and Liu, Dayiheng and Lin, Junyang and others},
  journal={arXiv preprint arXiv:2506.05176},
  year={2025}
}

@article{beltagy2020longformer,
  title={Longformer: The long-document transformer},
  author={Beltagy, Iz and Peters, Matthew E and Cohan, Arman},
  journal={arXiv preprint arXiv:2004.05150},
  year={2020}
}

@article{yang2025qwen3,
  title={Qwen3 technical report},
  author={Yang, An and Li, Anfeng and Yang, Baosong and Zhang, Beichen and Hui, Binyuan and Zheng, Bo and Yu, Bowen and Gao, Chang and Huang, Chengen and Lv, Chenxu and others},
  journal={arXiv preprint arXiv:2505.09388},
  year={2025}
}

@article{hu2022lora,
  title={Lora: Low-rank adaptation of large language models.},
  author={Hu, Edward J and Shen, Yelong and Wallis, Phillip and Allen-Zhu, Zeyuan and Li, Yuanzhi and Wang, Shean and Wang, Liang and Chen, Weizhu and others},
  journal={Iclr},
  volume={1},
  number={2},
  pages={3},
  year={2022}
}

\ifdefined\SOLNavWithSupplement
  % This file supports two build modes:
% 1. Compile Supplement.tex directly as a standalone document.
% 2. Include it from CameraReady2027.tex after the main-paper bibliography.
\ifdefined\SOLNavWithSupplement
  % The main manuscript already provides the document class and packages.
  \let\SOLNavEndSupplement\relax
\else
  \documentclass{article}
  \usepackage{amsmath,amssymb}
  \usepackage{graphicx}
  \usepackage{caption}
  \usepackage{subcaption}
  \usepackage{listings}
  \usepackage{xcolor}
  \usepackage{hyperref}
  \def\SOLNavEndSupplement{\end{document}}
\fi

% 代码样式配置，适配prompt文本展示
\lstset{
    basicstyle=\ttfamily\small,
    commentstyle=\color{gray},
    frame=single,
    breaklines=true,
    showstringspaces=false,
    numbers=none
}

\ifdefined\SOLNavWithSupplement
  % Keep the original single-column supplementary layout in the combined PDF.
  \onecolumn
\else
  \begin{document}
\fi

\appendix
\section{Supplementary Material}

\subsection{The Impact of Semantic Segmentation Performance on Building Structured Observation}

\begin{figure*}[h]
  \centering
  % \fbox{\rule{0pt}{0.5in} \rule{0.9\linewidth}{0pt}}
  \includegraphics[width=1.0\linewidth]{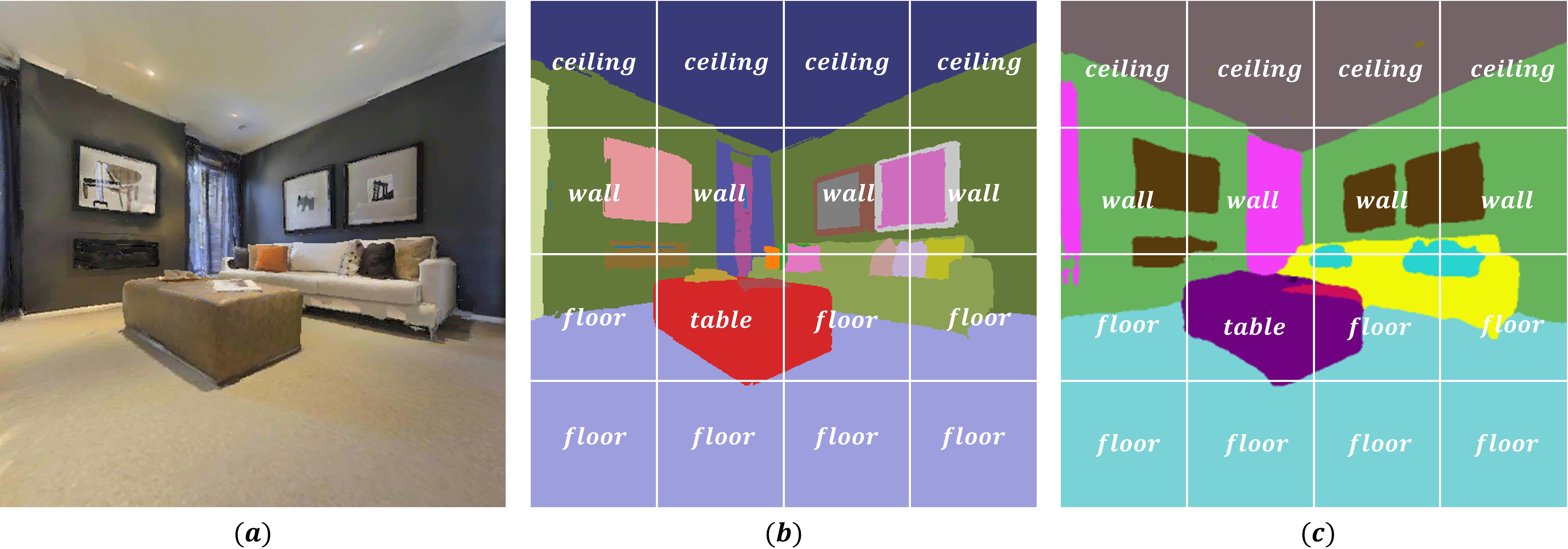}
  % \vspace{-10pt}
   \caption{\small Segmentation Impact on Structured Construction.}
   % \vspace{-12pt}
   \label{fig:seg}
\end{figure*}

Different segmentation models and the addition of noise do affect the semantic segmentation results. However, since we construct structured information by gridding the segmentation results, our approach inherently provides a filtering effect against segmentation errors. As shown in Figure \ref{fig:seg}, (b) and (c) represent good and poor segmentation results of (a), respectively. When we apply gridding and use the dominant pixel to represent each grid cell, the impact of segmentation errors on the constructed structured information is minimal. This makes our method highly robust to variations in semantic segmentation results. 
We also use random text masking and replacement during training to alleviate segmentation error propagation. 
Real-world annotation and SegFormer fine-tuning (FT) are optional for better deployment, not essential overhead. Zero-shot predictions from a pre-trained segmentation model already provide reasonable structured inputs. Moreover, FT of a segmentation model is far cheaper than training a VLN model from scratch, especially with foundation models or VLMs automating annotation.

\subsection{SOL-Nav Real-World Deployment Details}
\subsubsection{Hardware and Platform Setup}
\begin{figure}[htbp]
    \centering
    \includegraphics[width=1.0\linewidth]{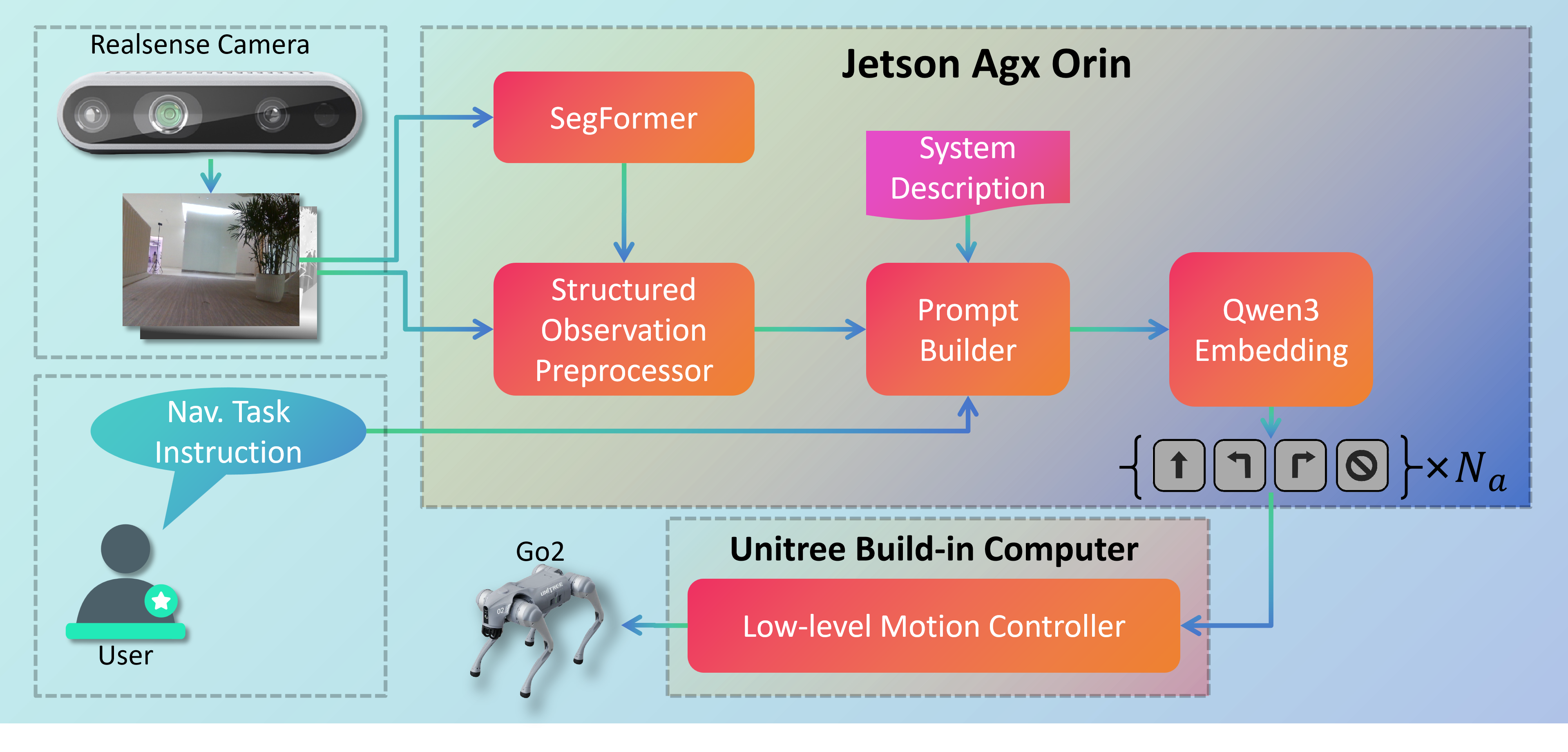}
    \caption{The SOL-Nav Deployment Pipeline.}
    \label{fig:deployment_pipeline}
\end{figure}

The SOL-Nav deployment pipeline is shown in Figure \ref{fig:deployment_pipeline}.
We conduct real-world deployment experiments using the \textbf{Unitree Go2} quadruped robot platform, with an NVIDIA Jetson AGX Orin as the edge computing unit and an Intel RealSense D435i RGB-D camera (fixed at \(640 \times 480\) resolution) for visual perception. The camera provides synchronized RGB and depth frames, which are fed into the SOL-Nav pipeline running on the Jetson Orin. Low-level motion control is handled by the Unitree Go2's built-in on-board computer, which executes high-level navigation commands from SOL-Nav.

\subsubsection{Inference Latency Breakdown on Jetson AGX Orin}
The end-to-end inference latency of SOL-Nav on the Jetson AGX Orin is approximately \(\textbf{0.8}\) seconds, decomposed into three core components:
\begin{itemize}
    \item SegFormer semantic segmentation: \(\sim 0.03\ \text{s}\)
    \item Structured Observation Preprocessor (SOP) + Prompt Builder (PB): \(\sim 0.22\ \text{s}\)
    \item Qwen3 Embedding (Qwen3-E) action prediction: \(\sim 0.53\ \text{s}\)
\end{itemize}

% Within the SOP+PB module, the latency breakdown is:
% \begin{enumerate}
%     \item Preprocessing (image loading + spatial alignment): \(0.0238\ \text{s}\)
%     \item Grid-based multi-modal feature extraction (total): \(0.1963\ \text{s}\)
%     \begin{itemize}
%         \item RGB feature extraction: \(0.0526\ \text{s}\)
%         \item Semantic feature extraction: \(0.0852\ \text{s}\)
%         \item Depth feature extraction: \(0.0580\ \text{s}\)
%     \end{itemize}
% \end{enumerate}
% The total latency of SOP+PB sums to \(0.2201\ \text{s}\), with semantic feature extraction being the dominant bottleneck.

\subsubsection{Action Chaining for Continuous Motion Execution}
To mitigate the inference latency and ensure continuous robot motion, we adopt a \textit{look-ahead action chaining strategy} based on predicted future action blocks. At each time step \(t\), the model predicts a sequence of \(N_a = 4\) future actions \(\{a_t^{(1)}, a_t^{(2)}, a_t^{(3)}, a_t^{(4)}\}\) given the current observation \(o_t\).

Let \(\mathcal{A}_t = \{a_t^{(1)}, a_t^{(2)}, \dots, a_t^{(N_a)}\}\) denote the action chunk predicted from observation \(o_t\). The executed action at step \(s\) (\(s = 1, 2, 3, \dots\)) is defined as:
\[
a_{\text{exec}}(s) =
\begin{cases}
a_{1}^{(1)}, & s = 1 \\
a_{s-1}^{(2)}, & s \geq 2
\end{cases}
\]
Where:
\begin{itemize}
    \item \(s = 1\): Execute the first action of the initial chunk \(\mathcal{A}_1\).
    \item \(s \geq 2\): Execute the second action of the chunk \(\mathcal{A}_{s-1}\) predicted from the previous observation \(o_{s-1}\).
\end{itemize}

This ensures that the robot always has a precomputed action to execute, eliminating idle time caused by inference latency and maintaining smooth navigation. A step-by-step execution flow is detailed as follows:
\begin{enumerate}
    \item \textbf{Initialization (Step 1):}
    Given the first observation \(o_1\), the model predicts an action chunk \(\mathcal{A}_1 = \{a_1^{(1)}, a_1^{(2)}, a_1^{(3)}, a_1^{(4)}\}\). The robot immediately executes \(a_1^{(1)}\).

    \item \textbf{Step 2:}
    After \(a_1^{(1)}\) completes, the robot captures a new observation \(o_2\) and initiates asynchronous inference to generate the next action chunk \(\mathcal{A}_2 = \{a_2^{(1)}, a_2^{(2)}, \dots, a_2^{(4)}\}\). Without waiting for the result of \(\mathcal{A}_2\), the robot executes \(a_1^{(2)}\) (the second action from the initial chunk \(\mathcal{A}_1\)).

    \item \textbf{Step 3 and beyond:}
    After \(a_1^{(2)}\) completes, the robot captures \(o_3\). By this time, the inference for \(\mathcal{A}_2\) has finished. The robot then executes \(a_2^{(2)}\) (the second action from chunk \(\mathcal{A}_2\)).
    This pattern repeats: for step \(s \geq 3\), the robot captures observation \(o_s\) and executes \(a_{s-1}^{(2)}\), which is the second action from the action chunk predicted by the previous observation \(o_{s-1}\).
\end{enumerate}
After the first step, the robot never waits for the inference result of the current observation, as it always executes the precomputed second action from the prior observation's chunk.

\subsection{An Example of the Structured Observation Prompt}
An example of the structured observation prompt generated from the R2R dataset is shown below:

\begin{lstlisting}
### System Description:
You are a robot that can turn left or right by a specific degree, move forward a certain distance, or stop. You must decide your next action based on the following sequence of time-stamped Observation Grids and the Task Instruction.

### Structured Observation:

[Time Step -18] Long Observation Grid:
[0,0]: depth=2.31, semantic=ceiling, color=light_gray; [0,1]: depth=2.31, semantic=ceiling, color=light_gray
[1,0]: depth=2.98, semantic=ceiling, color=gray; [1,1]: depth=2.98, semantic=ceiling, color=light_gray

[Time Step -17] Long Observation Grid:
[0,0]: depth=2.31, semantic=ceiling, color=light_gray; [0,1]: depth=2.31, semantic=ceiling, color=light_gray
[1,0]: depth=2.98, semantic=ceiling, color=light_gray; [1,1]: depth=2.98, semantic=ceiling, color=light_gray

[Time Step -16] Long Observation Grid:
[0,0]: depth=2.31, semantic=ceiling, color=light_gray; [0,1]: depth=2.31, semantic=ceiling, color=light_gray
[1,0]: depth=2.98, semantic=ceiling, color=light_gray; [1,1]: depth=2.98, semantic=ceiling, color=light_gray

[Time Step -15] Long Observation Grid:
[0,0]: depth=2.31, semantic=ceiling, color=light_gray; [0,1]: depth=2.31, semantic=ceiling, color=light_gray
[1,0]: depth=2.93, semantic=ceiling, color=light_gray; [1,1]: depth=2.97, semantic=ceiling, color=light_gray

[Time Step -14] Long Observation Grid:
[0,0]: depth=2.31, semantic=ceiling, color=light_gray; [0,1]: depth=2.31, semantic=ceiling, color=light_gray
[1,0]: depth=2.98, semantic=ceiling, color=light_gray; [1,1]: depth=2.98, semantic=ceiling, color=light_gray

[Time Step -13] Long Observation Grid:
[0,0]: depth=2.31, semantic=ceiling, color=light_gray; [0,1]: depth=2.31, semantic=ceiling, color=light_gray
[1,0]: depth=1.71, semantic=chair, color=light_gray; [1,1]: depth=2.97, semantic=ceiling, color=light_gray

[Time Step -12] Long Observation Grid:
[0,0]: depth=2.31, semantic=ceiling, color=light_gray; [0,1]: depth=2.31, semantic=ceiling, color=light_gray
[1,0]: depth=2.65, semantic=ceiling, color=light_gray; [1,1]: depth=2.98, semantic=ceiling, color=light_gray

[Time Step -11] Long Observation Grid:
[0,0]: depth=2.31, semantic=ceiling, color=light_gray; [0,1]: depth=2.31, semantic=ceiling, color=light_gray
[1,0]: depth=2.98, semantic=ceiling, color=gray; [1,1]: depth=2.98, semantic=ceiling, color=light_gray

[Time Step -10] Long Observation Grid:
[0,0]: depth=2.31, semantic=ceiling, color=light_gray; [0,1]: depth=2.31, semantic=ceiling, color=light_gray
[1,0]: depth=2.95, semantic=ceiling, color=yellow; [1,1]: depth=2.98, semantic=ceiling, color=gray

[Time Step -9] Long Observation Grid:
[0,0]: depth=2.31, semantic=ceiling, color=light_gray; [0,1]: depth=2.31, semantic=ceiling, color=light_gray
[1,0]: depth=2.88, semantic=ceiling, color=yellow; [1,1]: depth=2.93, semantic=ceiling, color=yellow

[Time Step -8] Long Observation Grid:
[0,0]: depth=2.32, semantic=ceiling, color=light_gray; [0,1]: depth=2.31, semantic=ceiling, color=light_gray
[1,0]: depth=2.77, semantic=wall, color=yellow; [1,1]: depth=2.84, semantic=wall, color=yellow

[Time Step -7] Long Observation Grid:
[0,0]: depth=2.32, semantic=ceiling, color=gray; [0,1]: depth=2.32, semantic=ceiling, color=light_gray
[1,0]: depth=2.59, semantic=wall, color=yellow; [1,1]: depth=2.70, semantic=wall, color=yellow

[Time Step -6] Long Observation Grid:
[0,0]: depth=2.31, semantic=ceiling, color=light_gray; [0,1]: depth=2.31, semantic=ceiling, color=light_gray
[1,0]: depth=2.93, semantic=ceiling, color=gray; [1,1]: depth=2.93, semantic=ceiling, color=yellow

[Time Step -5] Long Observation Grid:
[0,0]: depth=2.32, semantic=ceiling, color=light_gray; [0,1]: depth=2.32, semantic=ceiling, color=light_gray
[1,0]: depth=2.81, semantic=wall, color=gray; [1,1]: depth=2.81, semantic=wall, color=yellow

[Time Step -4] Long Observation Grid:
[0,0]: depth=2.32, semantic=ceiling, color=gray; [0,1]: depth=2.32, semantic=ceiling, color=light_gray
[1,0]: depth=2.61, semantic=wall, color=yellow; [1,1]: depth=2.61, semantic=wall, color=yellow

[Time Step -3] Long Observation Grid:
[0,0]: depth=2.27, semantic=ceiling, color=yellow; [0,1]: depth=2.27, semantic=ceiling, color=yellow
[1,0]: depth=2.40, semantic=wall, color=yellow; [1,1]: depth=2.44, semantic=wall, color=yellow

[Time Step -2] Short Observation Grid:
[0,0]: depth=2.11, semantic=wall, color=yellow; [0,1]: depth=2.11, semantic=wall, color=yellow; [0,2]: depth=2.11, semantic=wall, color=yellow; [0,3]: depth=2.48, semantic=wall, color=gray
[1,0]: depth=2.45, semantic=wall, color=light_gray; [1,1]: depth=2.77, semantic=wall, color=orange; [1,2]: depth=3.34, semantic=window, color=gray; [1,3]: depth=3.59, semantic=wall, color=yellow
[2,0]: depth=2.96, semantic=wall, color=gray; [2,1]: depth=3.81, semantic=wall, color=gray; [2,2]: depth=5.37, semantic=wall, color=gray; [2,3]: depth=4.62, semantic=wall, color=yellow
[3,0]: depth=2.95, semantic=wall, color=yellow; [3,1]: depth=3.66, semantic=door, color=light_gray; [3,2]: depth=4.54, semantic=wall, color=gray; [3,3]: depth=4.70, semantic=wall, color=yellow

[Time Step -1] Short Observation Grid:
[0,0]: depth=1.87, semantic=wall, color=yellow; [0,1]: depth=1.87, semantic=wall, color=yellow; [0,2]: depth=1.87, semantic=wall, color=gray; [0,3]: depth=3.09, semantic=wall, color=yellow
[1,0]: depth=2.75, semantic=wall, color=gray; [1,1]: depth=3.40, semantic=door, color=gray; [1,2]: depth=4.60, semantic=door, color=light_gray; [1,3]: depth=3.91, semantic=wall, color=yellow
[2,0]: depth=2.95, semantic=wall, color=gray; [2,1]: depth=3.80, semantic=wall, color=gray; [2,2]: depth=5.38, semantic=wall, color=gray; [2,3]: depth=4.01, semantic=wall, color=yellow
[3,0]: depth=2.70, semantic=door, color=gray; [3,1]: depth=3.13, semantic=door, color=light_gray; [3,2]: depth=4.33, semantic=wall, color=gray; [3,3]: depth=4.06, semantic=wall, color=yellow

[Time Step 0] Current Observation Grid:
[0,0]: depth=2.08, semantic=window, color=gray; [0,1]: depth=2.19, semantic=window, color=gray; [0,2]: depth=2.40, semantic=window, color=gray; [0,3]: depth=3.68, semantic=door, color=yellow; [0,4]: depth=5.76, semantic=door, color=light_gray; [0,5]: depth=5.44, semantic=ceiling, color=light_gray
[1,0]: depth=2.96, semantic=wall, color=gray; [1,1]: depth=3.80, semantic=door, color=gray; [1,2]: depth=4.39, semantic=door, color=yellow; [1,3]: depth=4.18, semantic=wall, color=yellow; [1,4]: depth=6.94, semantic=wall, color=yellow; [1,5]: depth=6.06, semantic=wall, color=yellow
[2,0]: depth=2.94, semantic=door, color=gray; [2,1]: depth=3.80, semantic=wall, color=gray; [2,2]: depth=4.46, semantic=wall, color=yellow; [2,3]: depth=4.27, semantic=wall, color=yellow; [2,4]: depth=7.11, semantic=wall, color=light_gray; [2,5]: depth=6.00, semantic=wall, color=yellow
[3,0]: depth=2.07, semantic=door, color=gray; [3,1]: depth=2.34, semantic=door, color=light_gray; [3,2]: depth=3.41, semantic=wall, color=yellow; [3,3]: depth=4.44, semantic=wall, color=light_gray; [3,4]: depth=7.21, semantic=wall, color=light_gray; [3,5]: depth=6.85, semantic=wall, color=light_gray
[4,0]: depth=1.02, semantic=cabinet, color=brown; [4,1]: depth=1.02, semantic=cabinet, color=light_gray; [4,2]: depth=1.39, semantic=cabinet, color=light_gray; [4,3]: depth=5.21, semantic=wall, color=gray; [4,4]: depth=11.21, semantic=wall, color=light_gray; [4,5]: depth=9.30, semantic=wall, color=yellow
[5,0]: depth=1.32, semantic=cabinet, color=dark_yellow; [5,1]: depth=1.02, semantic=cabinet, color=dark_yellow; [5,2]: depth=1.40, semantic=cabinet, color=light_gray; [5,3]: depth=4.09, semantic=wall, color=gray; [5,4]: depth=6.49, semantic=floor, color=light_gray; [5,5]: depth=6.14, semantic=floor, color=light_gray

### Task Instruction: 
Go around the right side of the center unit and stop by the right side doorway with the dining table and mirror in it.
\end{lstlisting}

% No-op when included; closes the standalone document otherwise.
\SOLNavEndSupplement

\fi

% ce requires a reproducibility checklist to be included in the paper.
% If so, you can uncomment the following line and ajust the path to include it.
% \input{ReproducibilityChecklist.tex}

\end{document}